\documentclass[11pt, a4paper, logo, copyright, nonumbering]{trillion}
\usepackage[authoryear, sort&compress, round]{natbib}
\usepackage{dblfloatfix}
\usepackage{ulem}
\usepackage{caption}
\usepackage{dramatist}
\usepackage{xspace}
\usepackage{pifont} 
\usepackage{booktabs}
\usepackage{multirow}
\usepackage{tcolorbox}
\usepackage{xltabular}
\usepackage{longtable}
\usepackage{hyperref}
\usepackage{amsfonts}
\usepackage{amsmath}
\usepackage{amssymb}
\usepackage{lineno}
\usepackage{multirow}
\usepackage{adjustbox}
\usepackage{graphicx}
\usepackage[bottom]{footmisc}

\usepackage{CJKutf8}
\usepackage{subfigure}
\usepackage{setspace}

\usepackage{dsfont}
\usepackage{array} 
\usepackage{tabularx} 
\usepackage{subfigure} 
\usepackage{xcolor} 

\usepackage{lipsum}  
\usepackage{multicol} 
\usepackage{xcolor}
\definecolor{mypurple}{HTML}{3C2C63}

\makeatletter
\def\@BTrule[#1]{%
  \ifx\longtable\undefined
    \let\@BTswitch\@BTnormal
  \else\ifx\hline\LT@hline
    \nobreak
    \let\@BTswitch\@BLTrule
  \else
     \let\@BTswitch\@BTnormal
  \fi\fi
  \global\@thisrulewidth=#1\relax
  \ifnum\@thisruleclass=\tw@\vskip\@aboverulesep\else
  \ifnum\@lastruleclass=\z@\vskip\@aboverulesep\else
  \ifnum\@lastruleclass=\@ne\vskip\doublerulesep\fi\fi\fi
  \@BTswitch}
\makeatother

\addto\extrasenglish{
}

 {\begin{list}{}%
         {\setlength{\leftmargin}{#1}}%
         \item[]%
 }
 {\end{list}}
 
\reportnumber{001} 

\renewcommand{\today}{}

\title{\centering Trillion 7B Technical Report}

\author[*]{
Trillion Labs
\\
\small
\texttt{research@trillionlabs.co}
}

\newcommand{\huggingface}{\raisebox{-1.5pt}{\includegraphics[height=1.05em]{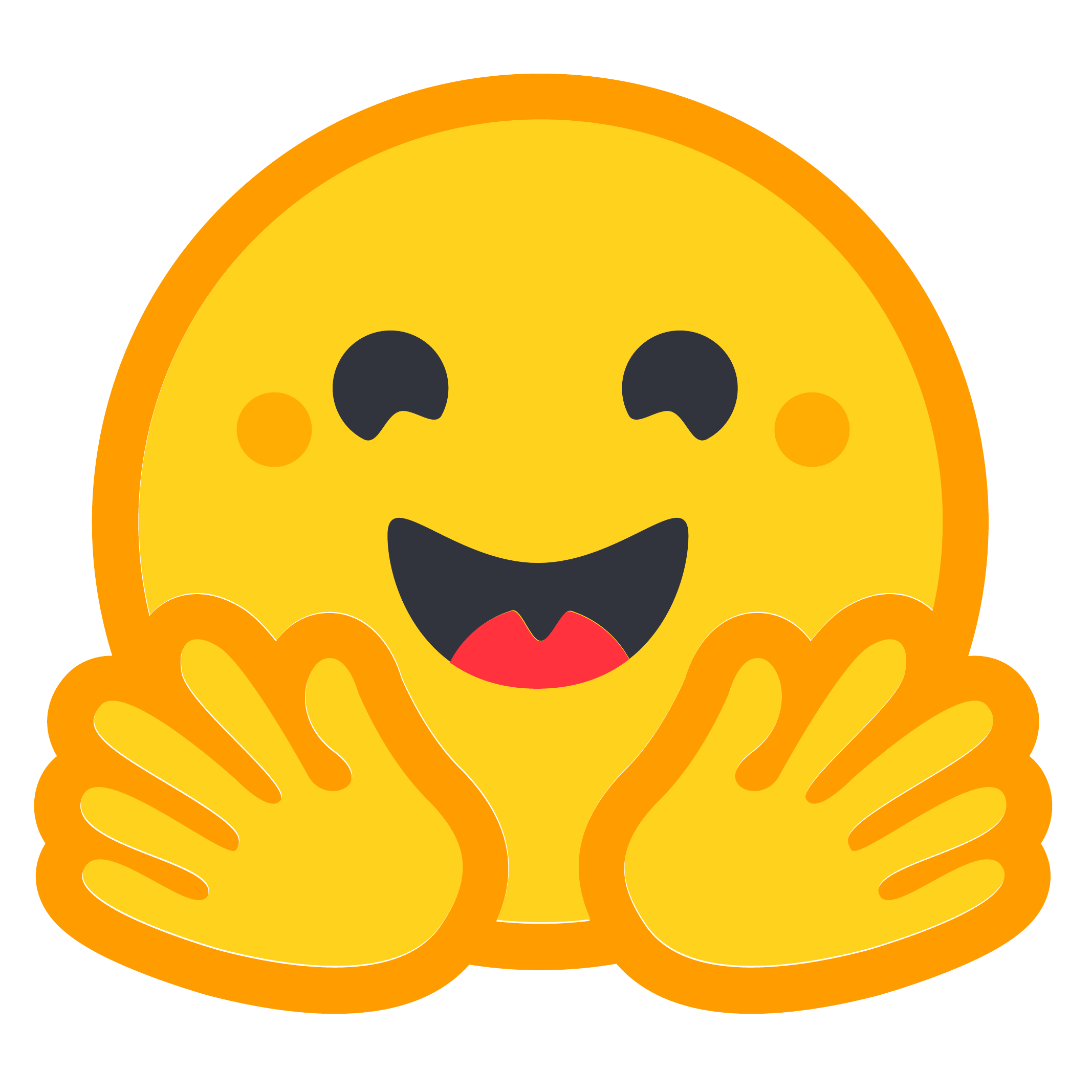}}\xspace}

\begin{abstract}

We introduce Trillion-7B, the most token-efficient Korean-centric multilingual LLM available. Our novel Cross-lingual Document Attention (XLDA) mechanism enables highly efficient and effective knowledge transfer from English to target languages like Korean and Japanese. Combined with optimized data mixtures, language-specific filtering, and tailored tokenizer construction, Trillion-7B achieves competitive performance while dedicating only 10\% of its 2T training tokens to multilingual data and requiring just 59.4K H100 GPU hours (\$148K) for full training. Comprehensive evaluations across 27 benchmarks in four languages demonstrate Trillion-7B's robust multilingual performance and exceptional cross-lingual consistency. 

\noindent\textbf{\huggingface Preview:} \href{https://huggingface.co/trillionlabs/Trillion-7B-preview}{\textcolor{mypurple}{\texttt{Trillion-7B}}} \quad \href{https://huggingface.co/trillionlabs/Trillion-LLaVA-7B}{\textcolor{mypurple}{\texttt{Trillion-LLaVA-7B}}}
\end{abstract}

\begin{document}
\begin{CJK*}{UTF8}{mj}

\maketitle

\begin{figure}[h]
\centering
\includegraphics[width=0.9\textwidth]{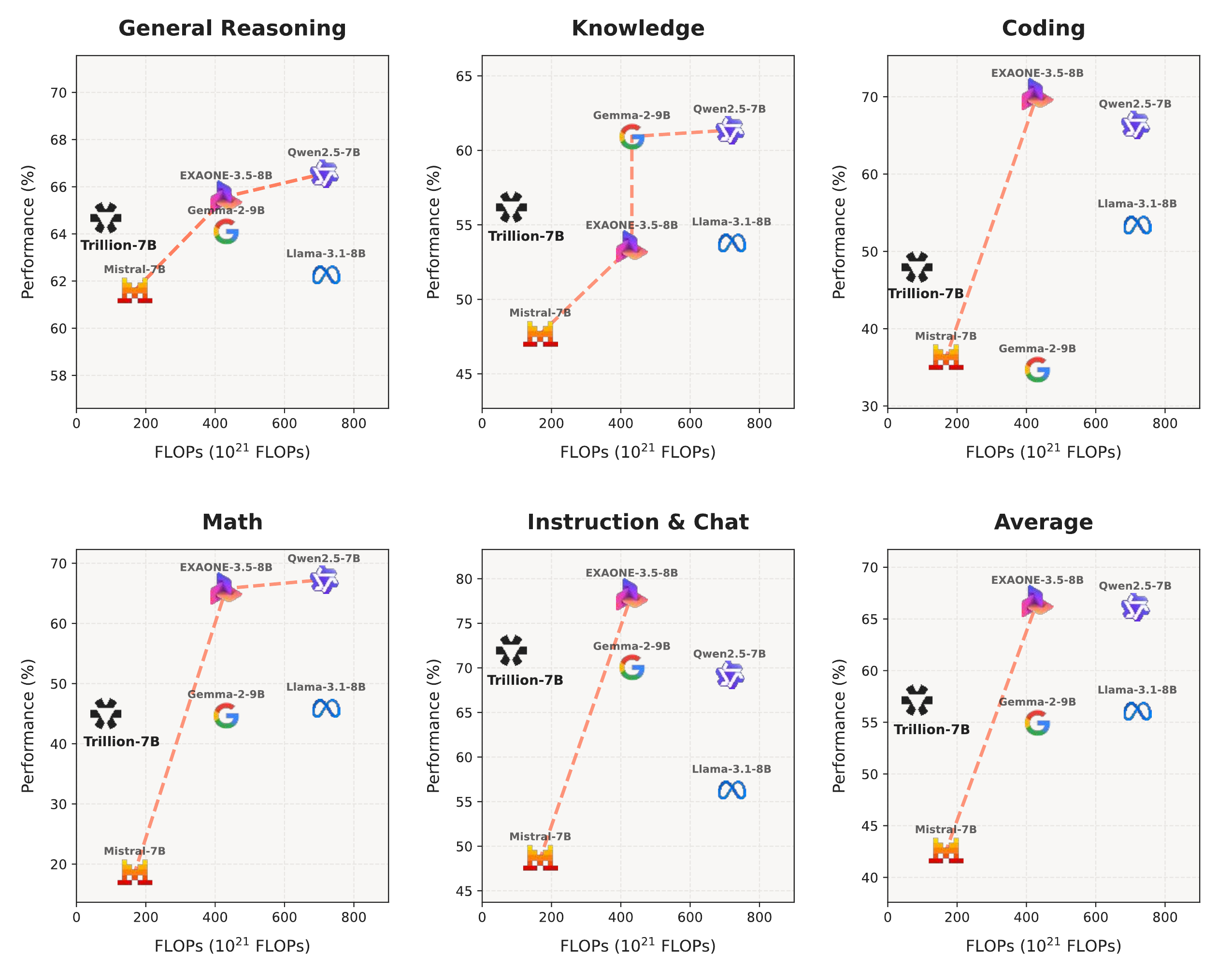}
\caption{
    \centering
    Trillion-7B significantly advances the Pareto-frontier across all aspects.
}
\label{fig:pareto_plot}
\end{figure}

\newpage
\begin{spacing}{0.9}
\tableofcontents
\end{spacing}
\newpage

\section{Introduction}

Recently, significant advancements have been made in developing massively multilingual large language models (LLMs) \citep{touvron2023llama2openfoundation,grattafiori2024llama3herdmodels,gemmateam2024gemma2improvingopen,qwen2025qwen25technicalreport,deepseekai2025deepseekv3technicalreport} expanding their coverage to hundreds of languages. However, substantial performance disparities still persist between high-resource languages like English and less resourced languages like Korean.

\begin{figure}[t]
    \centering
    \includegraphics[width=1\linewidth]{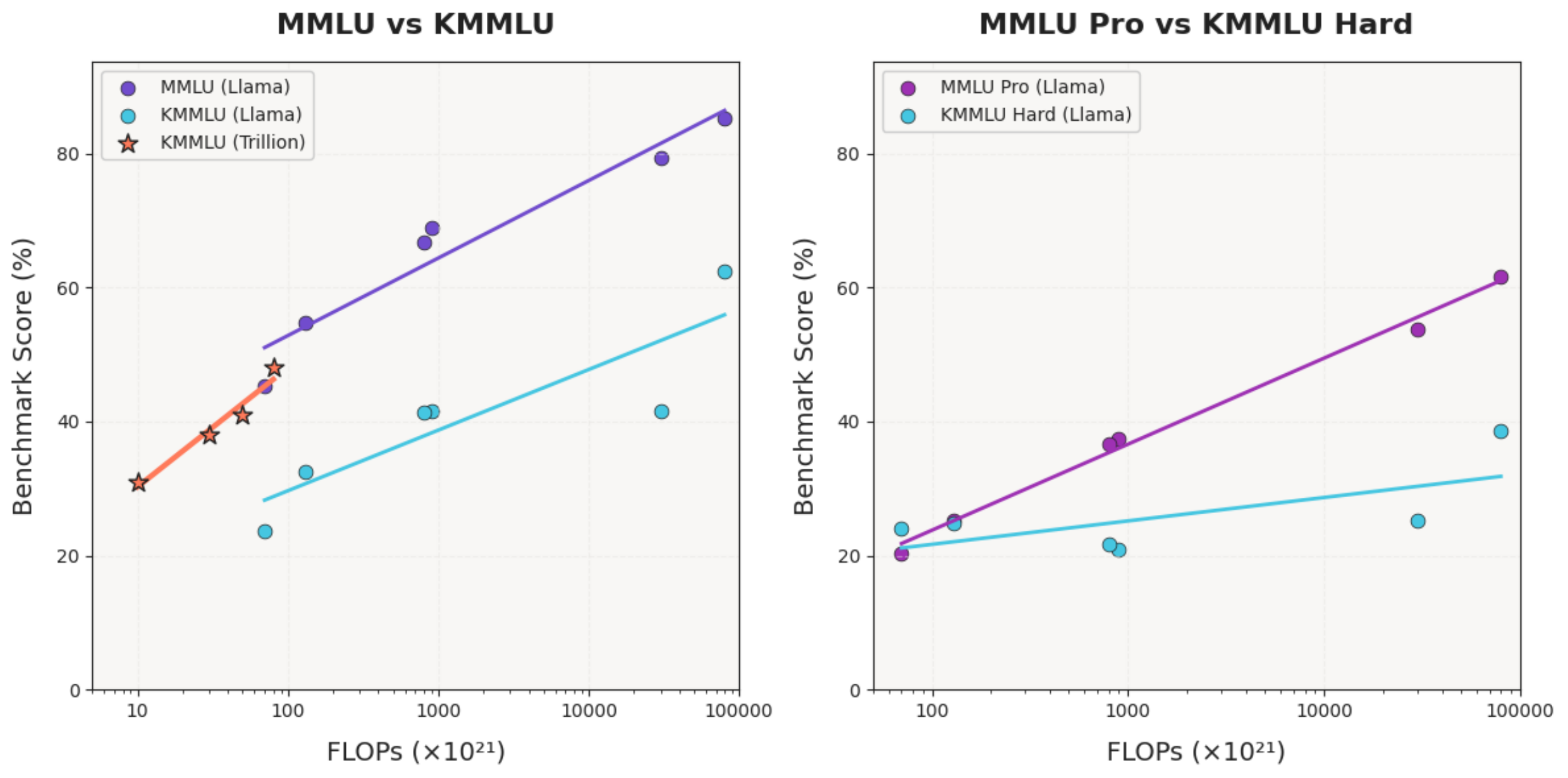}    
    \caption{
    \textbf{Discrepancy in scaling curves of Llama.} The above plots suggest that brute-force scaling (by Llama 2 \& 3) results in huge performance gaps between English and Korean, whereas Trillion-7B shows more desirable scaling laws for Korean performance closing the wide gap.}
    \label{fig:scaling}
\end{figure}

The fundamental reason for this discrepancy is data scale. Despite being spoken by over 81 million people worldwide, Korean content represents less than 2\% of English data volume in common web-crawled datasets like mC4 \citep{xue2021mt5massivelymultilingualpretrained}. This severe imbalance means that languages other than English and Chinese simply cannot follow the same scaling trajectory that has proven successful for extremely high-resource languages (Figure \ref{fig:scaling}). Addressing this limitation is critical to developing AI systems beneficial to all of humanity, as achieving true artificial general intelligence (AGI) inherently necessitates robust performance across diverse linguistic contexts.


In this technical report, we tackle the critical issue of data imbalance in multilingual training. We introduce Trillion-7B, a Korean-targeted multilingual model, which relies on our novel \textbf{Cross-lingual Document Attention (XLDA)} mechanism. XLDA functions as a form of architectural code-switching, efficiently transferring linguistic knowledge from a resource-rich language (like English) to a target less-resourced language (like Korean) without degrading English capabilities. Combined with other innovative training methodologies, we achieve exceptional multilingual performance, especially in Korean, while using significantly less multilingual tokens than prominent Korean-dominant approaches \citep{kim2021changeslargescalelanguagemodels,ko2023technicalreportpolyglotkoopensource,yoo2024hyperclovaxtechnicalreport,research2024exaone3078binstruction,research2024exaone35serieslarge}. Our comprehensive description of Trillion-7B focuses on its multilingual pre-training and post-training, detailing the architecture, methodologies, and transfer strategies that enable robust performance across diverse benchmarks.

Our main contributions are as follows:

\begin{itemize}
    \item \textbf{Cross-lingual Document Attention (XLDA)}: A novel mechanism designed explicitly for efficient cross-lingual knowledge integration.
    \item \textbf{Multilingual Token Efficiency Training}: Achieving competitive multilingual performance while dedicating approximately 10\% (< 220B tokens, with less than 180B in Korean) of the total 2T training tokens to multilingual data.
    \item \textbf{Complementary Technical Innovations}: Including optimized multilingual data mixtures, language-specific data filtering, customized tokenizer construction, and establishing empirical scaling laws to guide effective model scaling and training efficiency.
\end{itemize}

\begin{table}[h]
    \centering
    \setlength{\tabcolsep}{6pt}
    \begin{tabular}{@{}lccc@{}}
    \toprule
    \textbf{Training Costs} & \textbf{Pre-Training} & \textbf{Post-Training} & \textbf{Total} \\ \midrule
    in H100 GPU Hours       & 59K                   & 0.36K                  & 59.4K          \\
    in USD                  & \$147K                & \$896                  & \$148K         \\ \bottomrule
    \end{tabular}
    \caption{
    Training costs of Trillion 7B on 2T tokens, given H100 price as \$2.49 per GPU hour.
    }
    \label{tab:cost}
\end{table}

\begin{figure}[t]
    \centering
    \includegraphics[width=1.0\linewidth]{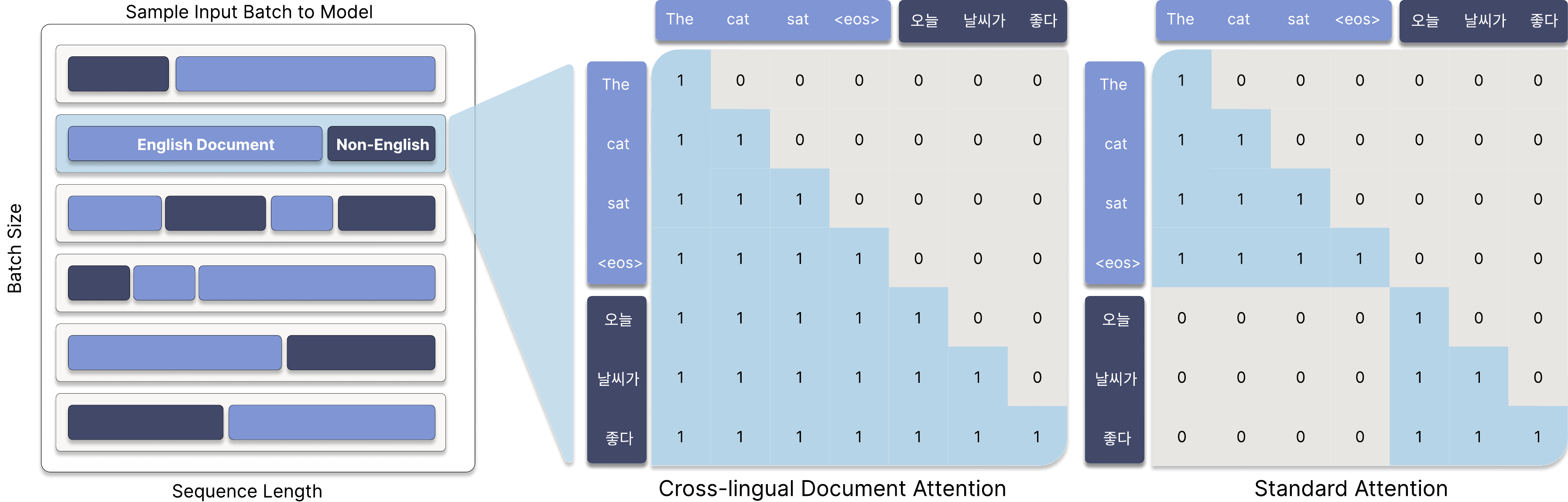}
    \caption{
    \textbf{Cross-Lingual Document Attention.} A multilingual batch (left) is packed so that each sequence contains contiguous spans from at least two languages (e.g. English + Korean). The XLDA mask (centre) keeps full self‑attention across language blocks (blue cells) while standard causal mask (right) blocks attention across document boundaries (grey cells). }
    \label{fig:xlda}
\end{figure}

\section{Cross-lingual Document Attention (XLDA) }
\label{sec:xlda}

During language model pre-training, it is common to pack several short documents into a single long sequence to maximize GPU utilization. Typically, training pipelines introduce segmentation masks at each document boundary to prevent tokens from attending to tokens in previous documents, thus avoiding cross-attention contamination \citep{Zhao_2024}.

However, in multilingual pre-training, this intra-document packing may unintentionally block beneficial cross-lingual correspondences that could naturally emerge when documents from different languages share the same context window. To address this limitation, we introduce a novel technique called Cross-Lingual Document Attention (XLDA), specifically designed to manage cross-lingual interactions effectively and improve token-level training efficiency.

\subsection{Strategic Batch-level Document Packing}

The first mechanism of XLDA focuses on how documents are packed into training batches. As illustrated in Figure~\ref{fig:xlda} (left), strategic batch-level document packing ensures that each sequence contains documents from multiple language sources—specifically, contiguous spans from at least two languages (English + non-English). We employ a controlled sampling strategy that combines documents from diverse languages at a predetermined rate, creating opportunities for cross-lingual learning. This intentional interleaving of linguistic content creates a rich training environment where the model can identify cross-lingual patterns and correspondences.

Formally, let $\mathcal{L} = \{l_1, l_2, \ldots, l_m\}$ be the set of $m$ languages in our corpus. For each training sequence $S$, we construct it as:

\begin{equation}
S = [d_1^{l_i}, d_2^{l_j}, \ldots, d_k^{l_n}]
\end{equation}

where $d_t^{l_x}$ represents the $t$-th document in language $l_x$. We \textit{enforce} the constraint that for each sequence $S$, there exist at least two documents $d_i^{l_p}$ and $d_j^{l_q}$ such that $l_p \neq l_q$, with a mixing probability $\rho$ that controls the likelihood of cross-lingual document adjacency. The sampling probability for constructing a batch with language $l_i$ is given by:

\begin{equation}
P(l_i) = \alpha \cdot \frac{|D_{l_i}|}{\sum_{j=1}^{m}|D_{l_j}|} + (1-\alpha) \cdot \beta_{l_i}
\end{equation}

where $|D_{l_i}|$ is the corpus size for language $l_i$, $\alpha$ is a temperature parameter that controls sampling smoothness, and $\beta_{l_i}$ is a language-specific upsampling factor for low-resource languages.

\subsection{Selective Attention Masking}

The second mechanism of XLDA introduces a novel approach to attention masking. As depicted in Figure~\ref{fig:xlda}, the XLDA mask keeps full self-attention across language blocks which allows tokens from different language documents to attend to each other. This contrasts with the standard causal mask that blocks attention across document boundaries. The visualization in Figure~\ref{fig:xlda} clearly illustrates how XLDA fundamentally differs from conventional masking approaches by permitting beneficial cross-document and cross-language attention flows. Note that this attention masking is effective because of the above enforced cross-lingual document packing.


\subsection{Design Rationale for XLDA}

\paragraph{Cross-lingual In-context Pretraining.}

We draw on insights from recent studies that proper document sequence packing can significantly enhance knowledge integration between the sequences being trained together in-context \citep{levine2022inductivebiasincontextlearning}. Through XLDA, our non-English data is consistently \textit{meta-learned} \citep{ortega2019metalearningsequentialstrategies, lampinen2024broaderspectrumincontextlearning} within a linguistic context that includes English, thereby strengthening the model's ability to generalize learned concepts across languages.

\paragraph{Synthetic Code Switching}

Recent works on code-switching \citep{yoo2024codeswitchingcurriculumlearningmultilingual, wang2025investigatingscalingcodeswitchingmultilingual} demonstrates substantial improvements in language alignment when models are exposed to mixed-language contexts. XLDA builds on this insight by enabling cross-document attention between languages during pretraining, creating natural opportunities for cross-lingual knowledge transfer without explicit code-switching data. 


\section{Pre-training}

\subsection{Pre-training Data}

The pretraining corpus for Trillion comprises approximately 2T tokens spanning English, multilingual, mathematical, and coding domains. The token distribution follows an 8.5 : 1 : 0.5 ratio (English : Korean : Other languages/Math/Code), creating an extremely English-predominant data distribution. This drastic imbalance in pretraining distribution is made viable specifically because of our XLDA mechanism, which enables effective cross-lingual transfer despite the asymmetric data representation. Within the portion other than English and Korean, we include Japanese, Chinese, Code, Math, and additional multilingual data. This language mixture encourages the model to develop core language-agnostic representations primarily in English \citep{zhao2024largelanguagemodelshandle, wendler2024llamasworkenglishlatent}, contrasting with an equally-balanced language distribution that are difficult to scale \citep{yoo2024hyperclovaxtechnicalreport, workshop2023bloom176bparameteropenaccessmultilingual} and can lead to negative interference between languages \citep{ye2023languageversatilistsvsspecialists}. We also emphasize substantial domain diversity within each language subset, driven by empirical observations indicating that data diversity significantly enhances model performance. This finding aligns with existing literature advocating for diverse pretraining data \citep{longpre2023pretrainersguidetrainingdata}.

\paragraph{Filtering} \label{data-filtering}
The scoring is performed using Qwen-72B-Instruct \citep{qwen2025qwen25technicalreport}, a state-of-the-art multilingual LLM that demonstrated strong correlation with GPT-4 \citep{openai2024gpt4technicalreport} in our evaluations. Due to computational constraints, we restrict direct scoring by Qwen-72B-Instruct to a carefully sampled subset of one million documents per language. We then utilize a portion of these scored documents to distill a smaller, multilingual embedding model, reserving the remainder as a held-out test set. By treating Qwen's assessments as gold-standard labels and binarizing scores at a threshold of 3, the distilled scoring model achieves an F1 score of 0.734 on our internal test set.
We use this distilled model for quality scoring across all pretraining data. For English, we retain the top 80\% of documents, while for other languages, we apply a much stricter filtering threshold, keeping only the top 50\% of multilingual data.

\subsection{Two-stage Pretraining}
Trillion 7B is pretrained in two stages, using a curriculum approach inspired by \citep{hu2024minicpmunveilingpotentialsmall, hagele2024scalinglawscomputeoptimaltraining, olmo20252olmo2furious}. We employ a warmup-stable-decay (WSD) scheduler that initially warms up to a high constant learning rate, followed by a learning rate decay. We define the \textit{annealing} phase as the point when the learning rate begins to decrease, and we adjust the data composition at this stage.
During the annealing phase, we enhance overall data quality and modify the mixture composition. For English-language data, we further curate documents by selecting only those within the top 20\% in terms of quality. For multilingual data, we apply an even stricter criterion, selecting only documents in the top 10\% quality threshold. This heightened quality reduces gradient noise, facilitating more effective optimization and enabling the model to better consolidate knowledge as it navigates through the loss valley \citep{hu2024minicpmunveilingpotentialsmall}.
Additionally, we significantly increase the proportion of multilingual data during the annealing phase, tripling its volume to further encourage cross-lingual knowledge transfer. We present ablative experiments on the effect of increasing quality and source distribution in Section \ref{annealing_exp}.

\subsection{Experiments on Scalable Training Recipe} \label{sec:scaling_training_recipe}

Efficiently validating training recipes and hyperparameters for large language models requires the strategic use of smaller-scale experiments. We employ a hybrid strategy combining established empirical scaling laws \citep{kaplan2020scalinglawsneurallanguage, hoffmann2022trainingcomputeoptimallargelanguage, deepseekai2024deepseekllmscalingopensource} with experimental evidence collected at smaller scales.

\begin{figure}[t]
    \centering
    \includegraphics[width=0.8\linewidth]{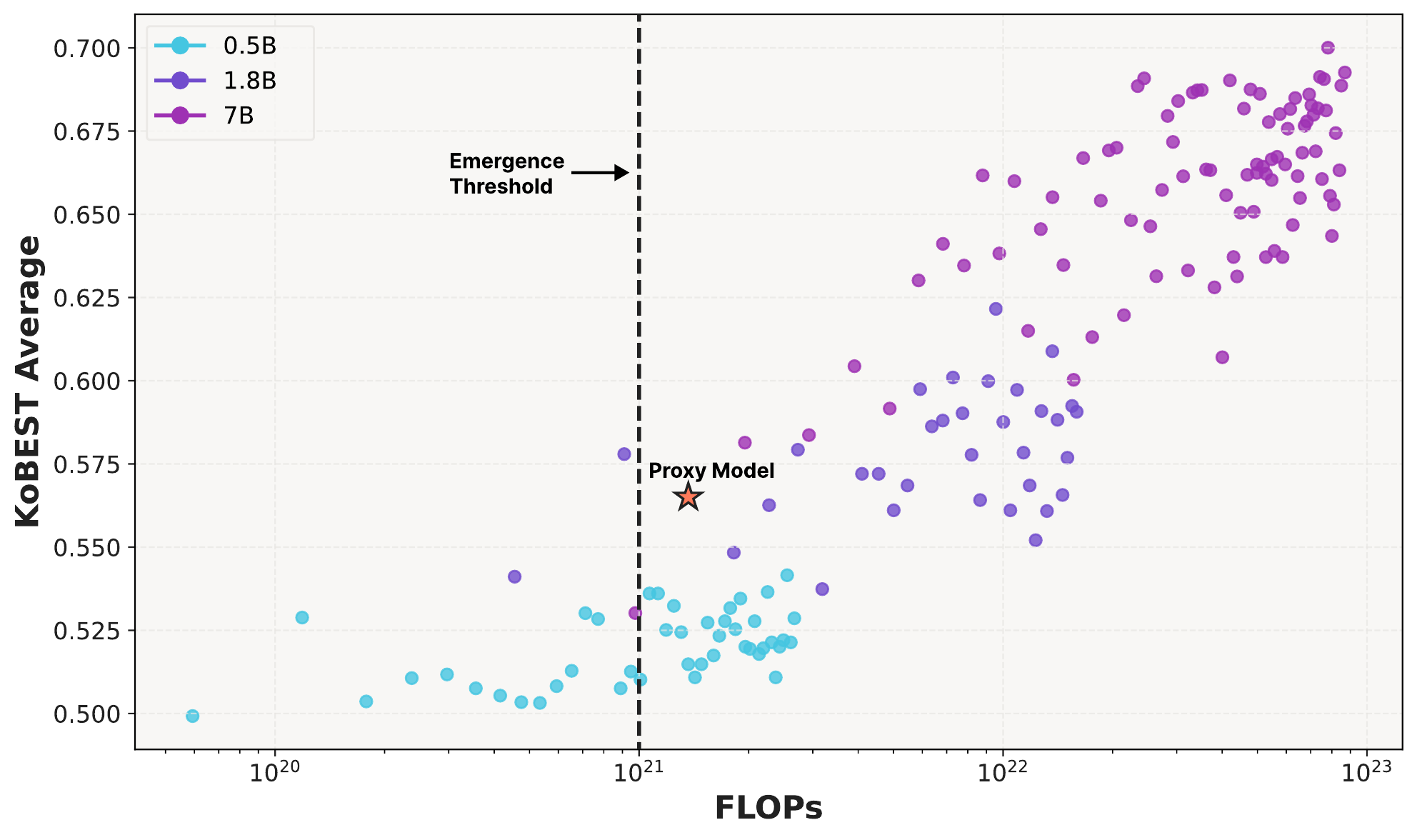}
    \centering
    \caption{\textbf{Proxy model and emergence point.} We trained 1.8B parameter models on approximately 100 billion tokens to serve as proxy models for determining optimal training configurations. This specific configuration, represented by a red star in the figure, identifies the most FLOP-efficient setting at which downstream task improvements become observable.}
    \label{fig:emergence}
\end{figure}

\paragraph{Emergence and Proxy Model} In selecting suitable configurations for small-scale experiments, we prioritize addressing the emergence phenomena observed in downstream tasks. These phenomena reflect non-linear improvements in downstream task performance after surpassing a critical threshold in validation loss \citep{du2025understandingemergentabilitieslanguage}.  We observe that this emergent threshold is tightly coupled with the model's emergence of in-context learning ability \citep{brown2020languagemodelsfewshotlearners}. Thus, we specifically curate benchmarks that emphasize few-shot learning along with a multilingual validation set consisting of approximately 100,000 documents. We find that a 1.8-billion-parameter model trained on approximately 100 billion tokens can serve as a good proxy for determining optimal training configurations as it can be efficiently trained while still observing emergence in downstream tasks improvements. Figure \ref{fig:emergence} illustrates the effectiveness of our proxy model on the KoBEST benchmark \citep{jang-etal-2022-kobest}. It can be observed that 0.5B scale does not provide a good test-bed due to large variance, while the emergence point is easily surpassed by the 1.8B model. In short, we identify the optimal training recipe in our 1.8 billion parameter setting and apply the known scaling laws to find the training recipe at 7 billion parameter scale.

\paragraph{Learning Rate} For critical hyperparameters such as learning rate,  we rely on results from Deepseek-V1 \cite{deepseekai2024deepseekllmscalingopensource} which found the law as $\mu \propto C^{-0.125}$, while holding constant less influential parameters. Applying this, we scale the optimal learning rate $\mu$ by 0.57 when scaling up. 

\paragraph{Vocabulary Size} For vocabulary size, we utilize the findings from \citep{tao2024scalinglawsvocabularylarger} which notes the power law relation of non-vocabulary size with compute $N_v \propto C^{0.42}$. Upon the experimental results on tokenizer vocabulary size (Section \ref{tokenizer_experiment}), applying this law suggests that we should increase the number of multilingual tokens by at least 6.3 times.

\subsection{Model Architecture}
Trillion is based on a Transformer decoder \citep{vaswani2023attentionneed} architecture with RoPE \citep{su2023roformerenhancedtransformerrotary} , SwiGLU \citep{shazeer2020gluvariantsimprovetransformer}, RMSNorm \citep{zhang2019rootmeansquarelayer}. It consists of 32 layers with a hidden size of 4096 and a feedforward dimension of 11008. We normalize the hidden representations before and after each transformer, attention, and feedforward layer. See Table \ref{tab:hyperparam} for details.

\begin{table}[h]
\centering
\renewcommand{\arraystretch}{1.2}
\begin{tabular}{l|c}
    \toprule
    \textbf{Model size} & \textbf{7B} \\
    \midrule
    Hidden dimension & 4096 \\
    Number of layers & 32 \\
    Feedforward dimension & 11,008 \\
    Number of heads & 32 \\
    Number of KV heads & 32 \\
    \midrule
    Activation Function & SwiGLU \\
    Max sequence length & 4096 \\
    Positional Embeddings & RoPE ($\theta=100,000$) \\
    Vocab size & 128,256 \\
    \bottomrule
\end{tabular}
\caption{\centering Model hyperparameters of Trillion}
\label{tab:hyperparam}
\end{table}

\paragraph{Multi-token Prediction} We adopt multi-token prediction (MTP) motivated by \citep{gloeckle2024betterfasterlarge,deepseekai2025deepseekv3technicalreport}. Along with the next token prediction (NTP) loss, we also predict the second next token by stacking a new transformer layer after the last layer for the prediction. The MTP loss is combined with the NTP loss through the hyperparameter $\alpha$. This MTP layer is discarded after pretraining and is not used during post-training.

\paragraph{Training and Hyperparameter Details}We utilize AdamW optimizer with learning rate of $2\times10^{-4}$, $(\beta_1, \beta_2) = (0.9, 0.95)$, weight decay of $0.1$, and 2000 warmup steps.  We use RoPE $\theta$ of 100,000. Batch size is slowly increased from 1M to 2M in the first 1T tokens as we observe that this leads to better training stability and faster emergence. A context length of 4096 is used throughout pretraining. MTP $\alpha$ of 0.2 is used. As mentioned above, we utilize the WSD learning rate scheduler, decaying the learning rate using inverse proportional decay function to 10\% ($2\times10^{-5}$) of its max learning rate  in the last 10\% of training. Hence, the model is trained at a high learning rate for 1.8T followed by 0.2T at the decaying rate. We also lower weight decay to 0.033 and $\alpha$ to 0.1 at the annealing stage. 

\subsection{Tokenizer}

Trillion 7B uses byte-level byte pair encoding (BPE) tokenizer \citep{Kida1999BytePE} \footnote{We utilize the \href{https://github.com/huggingface/tokenizers}{tokenizers} library for tokenizer training.}. Results from our preliminary experiments suggest that an excessively large vocabulary for non-English languages negatively impacts model performance due to sparse updates for infrequently appearing tokens, while too small vocabulary size reduces training efficiency by increasing token count. Further details are provided in Section \ref{tokenizer_experiment}.

\begin{figure}[t]
\centering
\includegraphics[width=0.9\linewidth]{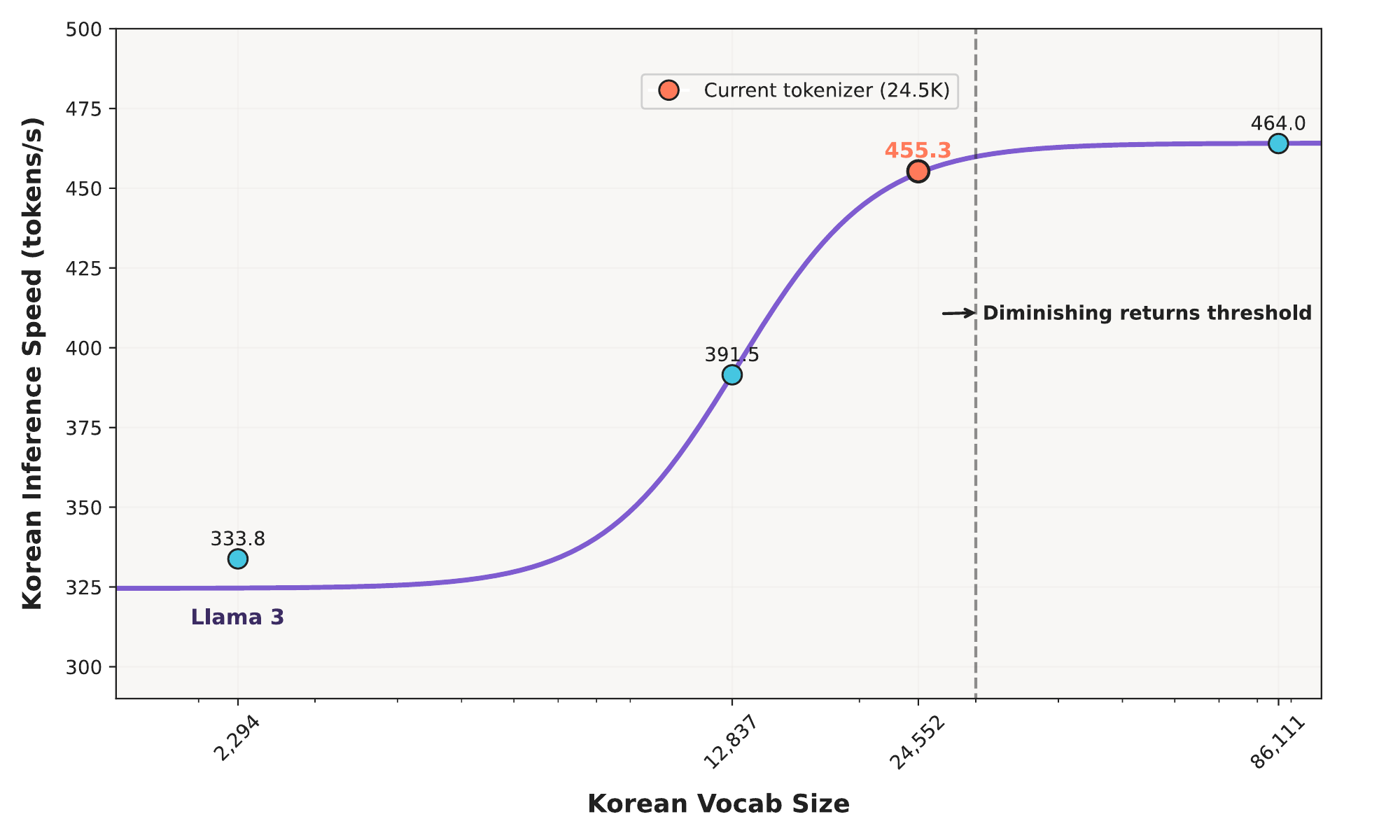}
\centering 
\caption{\textbf{Average Korean throughput measured on 1,000 selected Korean documents using vLLM.} We choose a Korean vocabulary size of 24,552 tokens, surpassing the scaling-law optimal size of around 13,000 tokens, yet still positioned just before the plateau in inference speed gains for Korean. This decision strategically balances theoretical optimality against practical improvements in inference speed.
} 
\label{fig:korean_tokens_improved_s_curve_v2}
\end{figure}

Our final tokenizer comprises 128,256 byte-level tokens, allocating approximately 100,000 tokens for English, 24,552 tokens for Korean, and the remaining tokens for other multilingual content. We accept a slight deviation from optimality (which is around 13,000 tokens suggested by the scaling law~\citep{tao2024scalinglawsvocabularylarger}) due to improvements in inference speed and consequently increased effective context length. In Figure \ref{fig:korean_tokens_improved_s_curve_v2}, it can be clearly seen that our chosen vocabulary is situated close to the inflection point of the inference speed plateau. Our adopted vocabulary size offers over 35\% increase in Korean inference speed while 13k only offers 11\% increase compared to Llama 3 tokenizer (leftmost point on the plot).

\subsection{Training Infrastructure}

Trillion 7B model was trained using 256 H100 GPUs, each with 80GB HBM3, using mixed precision training with BF16. We used fully sharded data parallelism \citep{zhao2023pytorchfsdpexperiencesscaling}. To reduce the high communication cost of the all-gather operation of FSDP, we sharded optimizer states only. We achieve HFU and MFU of 47.5\% and 42.5\% respectively. See Table \ref{tab:cost} training cost for Trillion-7B.

\subsection{Context Length Extension}
Longer context windows have become increasingly important for real-world applications in recent years. For context length extension, we employ a two-stage approach. On top of the pretrained model, we additionally train for 60B tokens with a context window of 32,768 tokens. We also extend the RoPE base frequency from 100,000 to 1,000,000 using the Adaptive Base Frequency technique \citep{xiong2023effectivelongcontextscalingfoundation}. Following \citep{gao2025trainlongcontextlanguagemodels}, we construct our training set with 60\% long context data and 40\% high-quality data with sequence lengths of at most 4,096 tokens. To support this long-context training, we utilize tensor parallelism \citep{shoeybi2020megatronlmtrainingmultibillionparameter} along with FSDP.

\section{Post-training}

For post-training, we closely follow the Tülu 3 framework consisting of Supervised Fine-Tuning (SFT), Direct Preference Optimization (DPO), and Reinforcement Learning with Verifiable Rewards (RLVR) \citep{lambert2025tulu3pushingfrontiers}. Since our work focuses primarily on multilingual pretraining, we adopt this established open-source post-training recipe to complete our model development pipeline.

\paragraph{Supervised Fine-Tuning (SFT)}
While primarily leveraging English data from Tulu 3~\citep{lambert2025tulu3pushingfrontiers}, we supplement our datasets with a smaller portion of open-sourced non-English prompt-response pairs in Korean, Japanese, and Chinese \citep{kopf2023openassistantconversationsdemocratizing}. We filter responses using the LLM-as-a-judge method \citep{zheng2023judgingllmasajudgemtbenchchatbot, kim2023prometheus, kim2024prometheus2, kim2024biggen}, employing Qwen-2.5-72B to score responses on a scale from 0 to 5. Only responses scoring above 3 are retained. The final SFT dataset contains approximately 800,000 prompt-response pairs. Importantly, we only apply the loss on responses. We leverage model merging \citep{wortsman2022modelsoupsaveragingweights} to combine three different checkpoints trained with different random seeds to produce the final SFT model.

\paragraph{Direct Preference Optimization (DPO)}
Following SFT, the model undergoes refinement via Direct Preference Optimization (DPO) \citep{rafailov2024directpreferenceoptimizationlanguage}, integrating around 200,000 pairs of preferred (winning) and non-preferred (losing) responses. We prompt our model to generate on-policy responses and pair these with T\"ulu 3 off-policy responses. The chosen-rejected pairs are selected using Qwen-2.5-72B as the reward model. To avoid unintended length biases, we carefully removed pairs with significantly different lengths.

\paragraph{Reinforcement Learning with Verifiable Rewards (RLVR)}
Finally, Trillion 7B is fine-tuned on a targeted set of 10,000 prompts, predominantly mathematical questions from the training sets of GSM8k \citep{cobbe2021trainingverifierssolvemath} and MATH \citep{hendrycks2021measuringmathematicalproblemsolving}. This stage uses reinforcement learning with verifiable feedback via Group Relative Policy Optimization (GRPO) \citep{shao2024deepseekmathpushinglimitsmathematical}. We carefully design reward functions for both mathematical reasoning and instruction-following abilities. We found that using problems with appropriate difficulty levels and implementing the right system prompt are crucial factors in optimizing performance. 

\section{Evaluation}

This section presents the evaluation setup and results of Trillion language models, compared with open language models of comparable size across benchmarks. For details on the baseline models, see Appendix~\ref{baseline_models}. Plotting Trillion and the baselines in terms of compute against the multilingual benchmark scores, we achieve a strong performance, see Figure \ref{fig:pareto_plot}. 

\subsection{Benchmarks}

We evaluated our models on 27 benchmarks across four languages, categorized into General Reasoning, Knowledge, Mathematical Reasoning, Coding, and Instruction-Following tasks. Table~\ref{tab:benchmarks} summarizes these benchmarks, and full details are in Appendix~\ref{app:full_details}. We provide detailed information about the prompts and example instances used in our evaluation in Appendix \ref{sec:eval_prompts}.


\begin{table}[ht]
    \centering
    \resizebox{\textwidth}{!}{%
\begin{tabular}{@{}lllll@{}}
\toprule
\textbf{Category}                               & \textbf{Name}                                  & \textbf{Language} & \textbf{Prompting}    & \textbf{Metric}         \\ \midrule
\multirow{10}{*}{\textbf{General Reasoning}}    & HellaSwag \cite{zellers2019hellaswag}                     & English  & 0-shot       & accuracy       \\
                                                & TruthfulQA\_mc1 \cite{lin2022truthfulqa}                  & English  & 6-shot       & accuracy       \\
                                                & TruthfulQA\_mc2 \cite{lin2022truthfulqa}                  & English  & 6-shot       & accuracy       \\
                                                & ARC:C \cite{allenai:arc}                                  & English  & 0-shot       & accuracy       \\
                                                & HAERAE \cite{son2024haerae}                               & Korean   & 3-shot       & accuracy       \\
                                                & KoBEST \cite{jang-etal-2022-kobest}                       & Korean   & 5-shot       & accuracy       \\
                                                & BBH \cite{suzgun2022challenging}                          & English  & 3-shot, CoT  & accuracy       \\
                                                & xwinograd\_en \cite{muennighoff2023xwinograd}             & English  & 0-shot       & accuracy       \\
                                                & xwinograd\_jp \cite{muennighoff2023xwinograd}             & Japanese & 0-shot       & accuracy       \\
                                                & xwinograd\_zh \cite{muennighoff2023xwinograd}             & Chinese  & 0-shot       & accuracy       \\ \midrule
\multirow{7}{*}{\textbf{Knowledge}}             & KMMLU \cite{son2024kmmlu}                                 & Korean   & 5-shot       & accuracy       \\
                                                & MMLU \cite{hendrycks2021mmlu}                             & English  & 5-shot       & accuracy       \\
                                                & GMMLU-en \cite{singh2024globalmmlu}                       & English  & 5-shot       & accuracy       \\
                                                & GMMLU-ko \cite{singh2024globalmmlu}                       & Korean   & 5-shot       & accuracy       \\
                                                & GMMLU-ja \cite{singh2024globalmmlu}                       & Japanese & 5-shot       & accuracy       \\
                                                & GMMLU-zh \cite{singh2024globalmmlu}                       & Chinese  & 5-shot       & accuracy       \\
                                                & GPQA \cite{rein2023gpqa}                                  & English  & 4-shot       & accuracy       \\ \midrule
\multirow{3}{*}{\textbf{Math}}                  & GSM8k \cite{cobbe2021gsm8k}                               & English  & 0-shot, CoT  & exact-match    \\
                                                & MATH \cite{hendrycksmath2021}                             & English  & 0-shot, CoT  & exact-match    \\
                                                & HRM8k \cite{ko2025hrm8k}                                  & Korean   & 0-shot, CoT  & exact-match    \\ \midrule
\multirow{2}{*}{\textbf{Coding}}                & MBPP \cite{austin2021mbpp}                                & English  & 0-shot, CoT  & pass@1         \\
                                                & HumanEval \cite{chen2021humaneval}                        & English  & 0-shot, CoT  & pass@1         \\ \midrule
\multirow{5}{*}{\textbf{Instruction-following}} & IFEval \cite{zhou2023ifeval}                              & English  & 0-shot       & strict-average \\
                                                & Ko-IFEval$^*$                                             & Korean   & 0-shot       & strict-average \\
                                                & MT-Bench \cite{zheng2023judgingllmasajudgemtbenchchatbot} & English  & LLM-as-judge & LLM score      \\
                                                & KO-MT-Bench \cite{KoMT-Bench}                             & Korean   & LLM-as-judge & LLM score      \\
                                                & LogicKor \cite{logickor}                                  & Korean   & LLM-as-judge & LLM score      \\ \bottomrule 
\end{tabular}
    }
\begin{flushleft}
\footnotesize $^*$Our in-house instruction following evaluation set in Korean.
\end{flushleft}
\caption{\centering Summary of evaluation benchmarks}
\label{tab:benchmarks}
\end{table}

\subsection{Evaluation Results}

\begin{table}[ht]
    \small  
    \resizebox{\textwidth}{!}{%
    \begin{tabular}{@{}lrrrrrr@{}}
    \toprule
    \textbf{Category} & \textbf{Trillion-7B} & \textbf{Qwen2.5-7B} & \textbf{EXAONE-3.5-8B} & \textbf{Gemma-2-9B} & \textbf{Llama-3.1-8B} & \textbf{Mistral-7B} \\ \midrule
    \textbf{Training Tokens} & 2T             & 17T            & 9T             & 8T    & 15T   & 4T    \\ \midrule
    General Reasoning        & 64.67          & \textbf{66.56} & 65.58          & 64.10 & 62.27 & 61.60 \\
    Knowledge                & 56.17          & \textbf{61.36} & 53.56          & 60.93 & 53.79 & 47.68 \\
    Coding                   & 47.94          & 66.35          & \textbf{70.33} & 34.69 & 53.44 & 36.30 \\
    Math                     & 45.02          & \textbf{67.29} & 65.82          & 43.60 & 46.68 & 18.65 \\
    Instruction \& Chat      & 71.96          & 69.20          & \textbf{78.27} & 70.06 & 56.28 & 48.71 \\ \midrule
    Average (Rank)    & 57.15 (3)      & 66.15 (2)      & \textbf{66.71 (1)}      & 54.68 (4)      & 54.49 (5)      & 42.59 (6)      \\ \bottomrule    
    \end{tabular}
}
\caption{\centering Model Performance Summary by Category (\%). Macro-averaged.}
\label{tab:performance-summary}
\end{table}

\paragraph{Benchmark Results}

Table~\ref{tab:performance-summary} presents the performance of Trillion models compared to baselines across all benchmarks. Our models demonstrated strong capabilities across various languages, with particularly impressive results on multilingual chat and instruction following.

\section{Ablations}

We present various ablative experiments illustrating the effectiveness of Trillion 7B's training recipe. Our ablations specifically focus on four languages: English, Korean, Japanese, and Chinese.

\subsection{Data Composition and Quality}

\begin{table}[ht]
\centering
\begin{tabular}{@{}lccccc@{}}
\toprule
         &                    & \multicolumn{2}{c}{\textbf{KoBEST}} & \multicolumn{2}{c}{\textbf{XWino}} \\ \cmidrule(l){3-6} 
         & \textbf{HellaSwag} & \textbf{COPA}  & \textbf{HellaSwag} & \textbf{ZH}      & \textbf{JP}     \\ \midrule
No QF & 0.3990 & 0.616 & 0.370 & 0.6429 & 0.5892 \\
QF-Top50 & \textbf{0.4399}    & \textbf{0.671} & \textbf{0.392}     & \textbf{0.6448}  & \textbf{0.6173} \\ \bottomrule
\end{tabular}
\caption{\centering Performance comparison with and without quality filtering on Korean data.}
\label{tab:korean_qf_results}
\end{table}

\paragraph{Pre-training} We present the effectiveness of our quality filtering approach for the pretraining phase using 1.8 billion parameter model on 100 billion tokens. Specifically, we improved the quality of Korean documents by selecting the 50 percentile (\textit{QF-Top50)} while leaving other languages unchanged and compare this to the baseline with no quality filtering (\textit{No QF}). Table \ref{tab:korean_qf_results} presents the results, showing improvement across all languages despite only enhancing the quality of Korean data. 

\paragraph{Annealing} \label{annealing_exp} We conducted ablative experiments to evaluate the significance of the annealing stage by altering the composition and quality of training data over an additional 20 billion tokens, starting from a 7B model trained up to 1.8 trillion tokens at a high learning rate. Using a model trained with consistently high learning rate as our baseline (\textit{High Learning Rate}), we experimented with three variants: \textit{Anneal} (decayed learning rate with original data mixture), \textit{Anneal+Quality} (decayed learning rate with high-quality filtered data), and \textit{Anneal+Quality+Composition} (decayed learning rate with high-quality filtered data and up-sampled multilingual, math and code composition). Table \ref{tab:annealing_gmmlu_results} presents the Global MMLU results for English, Korean, Japanese, and Chinese. The findings underscore the critical role of quality filtering, which significantly enhances performance across all languages. An intriguing observation is the improvement in English performance under the \textit{Anneal+Quality+Composition} condition, despite a reduction in English data volume, highlighting the cross-lingual knowledge bridging effects discussed in Section \ref{sec:xlda}. Another contributing factor for this improvement is the up-sampling of math and code data, which is aligned with previously reported empirical results on the effects of such data on downstream performance \citep{petty2025doescodepretrainingaffect, blakeney2024doesdatasparkjoy}.

\begin{table}[h!]
\centering
\begin{tabular}{@{}lcccc@{}}
\toprule
                   & \multicolumn{4}{c}{\textbf{GMMLU}} \\ \cmidrule(l){2-5} 
\textbf{Checkpoint}        & \textbf{EN}     & \textbf{KO}     & \textbf{JA}   & \textbf{ZH}    \\ \midrule
High Learning Rate & 0.530   & 0.44   & 0.4225 & 0.4275 \\
Anneal             & 0.5675  & 0.445  & 0.435  & 0.4875 \\
Anneal+Quality     & 0.61    & 0.4550 & 0.4675 & 0.4977 \\
Anneal+Quality+Composition & \textbf{0.6325} & \textbf{0.4875} & \textbf{0.50} & \textbf{0.505} \\ \bottomrule
\end{tabular}
\vspace{0.5em}
\centering 
\caption{Checkpoint performance on GMMLU for different annealing strategies. GMMLU results are shown for English (EN), Korean (KO), Chinese (ZH), and Japanese (JA). Bold indicates the best scores per column.}
\label{tab:annealing_gmmlu_results}
\end{table}


\subsection{Impact of Data Diversity}

We further investigated the impact of multilingual data diversity by combining the non-English multilingual FineWeb dataset \citep{penedo2024finewebdatasetsdecantingweb} with our in-house Korean pretraining data. Three models at the 1.8B scale were trained: one baseline model using only FineWeb as the multilingual dataset, and two variants with incrementally increased diversity from our in-house dataset. Deduplication was applied with FineWeb to ensure results were not confounded by duplicated texts. Results shown in Table \ref{tab:diversity_comp} reveal that enhancing diversity in the Korean dataset consistently boosts performance in Korean-specific benchmarks. Furthermore, the results confirm effective multilingual transfer, as evidenced by performance gains in other languages. Notably, mutual knowledge transfer improvements are observed among English, Japanese, and Korean benchmarks.

\begin{table}[ht]
\centering
\begin{tabular}{@{}lccccc@{}}
\toprule
 & \multirow{2}{*}{\textbf{HellaSwag}} & \multicolumn{2}{c}{\textbf{XWinogrande}} & \multirow{2}{*}{\textbf{KoBEST}} & \multirow{2}{*}{\textbf{Average}} \\ \cmidrule(lr){3-4}
                     &                & \textbf{JA}    & \textbf{ZH}    &                &                \\ \midrule
FineWeb              & 0.544          & 0.696          & 0.698          & 0.516          & 0.614          \\
FineWeb + In-House 1 & \textbf{0.603} & 0.705          & \textbf{0.700} & 0.549          & 0.639          \\
FineWeb + In-House 2 & 0.602          & \textbf{0.711} & 0.692          & \textbf{0.635} & \textbf{0.660} \\ \bottomrule
\end{tabular}
\caption{\centering Korean performance comparison by increasing data diversity}
\label{tab:diversity_comp}
\end{table}



\subsection{Vocabulary Size} \label{tokenizer_experiment}

\begin{figure}[t]
    \centering
    \includegraphics[width=0.6\linewidth]{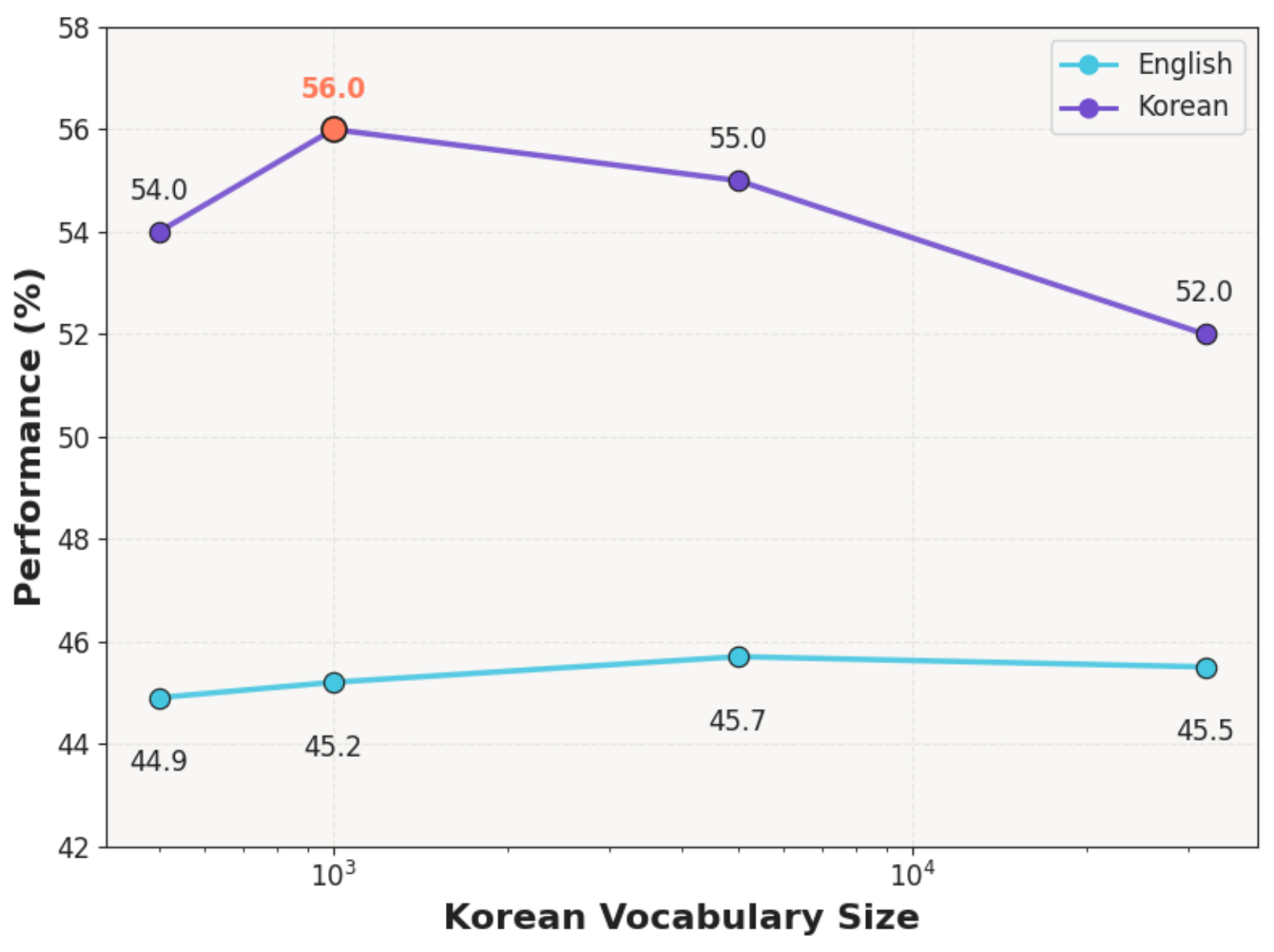} 
    \caption{Tokenizer ablation experiments on Korean vocabulary size with 1.8B model trained on 100 billion tokens. We report KoBEST for Korean and HellaSwag for English.}
    \label{fig:tokenizer_exp}
\end{figure}

We perform an ablative experiment to find the optimal vocabulary size for Korean. We show that using a smaller vocabulary can actually be beneficial for Koreans, contrary to the established knowledge that increasing language-specific vocabulary can lead to an improved performance \citep{seo2025doeslanguagespecifictokenizeraffect}. With 1.8B model trained with 100 billion tokens, we conduct a controlled experiment by ablating the number of Korean tokens while keeping other non-Korean entries fixed in the tokenizer. The resulting tokenizers are subsets of their larger counterparts. See Figure \ref{fig:tokenizer_exp} for results. While we see a relatively stable performance in English, we see a drastic performance difference in Korean, observing that our optimal Korean vocabulary size is situated between 1500 and 5000 tokens, which is further adjusted by the vocabulary scaling laws suggested from \citep{tao2024scalinglawsvocabularylarger}.

\section{Analysis}

\subsection{Cross-lingual Consistency}
We demonstrate our model builds robust multilingual representations capable of effective cross-lingual generalization by assessing prediction consistency across languages. We focus on whether English knowledge transfers properly to Korean using parallel question sets from Global-MMLU \citep{singh2024globalmmlu}, translated by language experts.
We define three metrics based on prediction consistency. The primary metric measures whether correct English predictions lead to correct Korean predictions:
\begin{equation}
E(T) \rightarrow K(T) = \mathbb{E}_{i \in Q} \left[ \text{Model}(Q_i^{ko}, A_i^{ko}) = 1 \mid \text{Model}(Q_i^{en}, A_i^{en}) = 1 \right]
\end{equation}
Here, \( Q_i^{en} \) and \( A_i^{en} \) represent the English question and gold answer, while \( \text{Model}(Q_i^{en}, A_i^{en}) = 1 \) indicates correct prediction. Similarly for Korean \( Q_i^{ko} \) and \( A_i^{ko} \).
We also assess whether incorrect predictions in one language can revert in another:
\begin{equation}
E(F) \rightarrow K(T) = \mathbb{E}_{i \in Q} \left[ \text{Model}(Q_i^{ko}, A_i^{ko}) = 1 \mid \text{Model}(Q_i^{en}, A_i^{en}) = 0 \right]
\end{equation}
This metric reveals consistency in knowledge representation across languages. Table \ref{tab:transformation_consistency_results} shows Trillion achieves superior consistency across all measures compared to leading multilingual models, confirming its robust cross-lingual generalization capability.

\begin{table}[ht]
\begin{tabular}{@{}lcccc@{}}
\toprule
Model & \textbf{Tokens} & \textbf{E(T)→K(T)} $\uparrow$ & \textbf{K(F)→E(T)} $\downarrow$ & \textbf{E(F)→K(T)} $\downarrow$ \\ \midrule
Trillion 7B     & 2T  & \textbf{77.5}\% & \textbf{29.8}\% & \textbf{18.73}\% \\
Llama 3.1 7B    & 15T & 67.7\%          & 42.0\%          & 18.74\%          \\
Exaone 3.5 7.8B & 9T  & 76.4\%          & 38.5\%          & 21.21\%          \\
Qwen 2.5 7B     & 17T & 77.4\%          & 42.8\%          & 19.83\%          \\ \bottomrule
\end{tabular}
\centering
\caption{\centering Consistency comparison between Korean and English.}
\label{tab:transformation_consistency_results}
\end{table}

\subsection{Generalization to Vision}
We demonstrate Trillion 7B's robust multilingual representation enables effective transfer to vision modalities. Following LLaVA's approach \citep{liu2023visualinstructiontuning}, we finetune our model into a Vision Language Model (VLM) using identical training framework for controlled comparison. Despite training exclusively on English vision-language instruction pairs, the model shows strong performance on Korean visual reasoning tasks.
Table~\ref{tab:cross-lingual-performance} compares Trillion-LLaVA's performance on English and Korean VLM benchmarks \citep{ju2024varcovision} against comparable models.

\begin{table}[htbp]
\centering
\small 
{%
\begin{tabular}{l*{4}{cc}c} 
\toprule
& \multicolumn{2}{c}{\textbf{MMBENCH}} & \multicolumn{2}{c}{\textbf{SEED-I}} & \multicolumn{2}{c}{\textbf{MMStar}} & \multicolumn{1}{c}{\textbf{K-DTCB}} \\
\cmidrule(lr){2-3} \cmidrule(lr){4-5} \cmidrule(lr){6-7} \cmidrule(lr){8-8}
\textbf{Model} & En & Ko & En & Ko & En & Ko & Ko \\
\midrule
\textbf{Llava-1.5-Vicuna-7B} & 0.64 & 0.43 & 0.66 & 0.52 & 0.34 & 0.33 & 0.30 \\
\textbf{Llava-1.6-Mistral-7B} & \textbf{0.68} & 0.49 & \textbf{0.72} & 0.61 & 0.36 & 0.33 & 0.30 \\
\textbf{Trillion-LLaVA-7B} & 0.66 & \textbf{0.61} & 0.68 & \textbf{0.66} & \textbf{0.37} & \textbf{0.37} & \textbf{0.33} \\
\bottomrule
\end{tabular}%
}
\centering
\caption{Performance comparison (English, Korean) across different vision-language models. Although Trillion-LLaVA-7B was trained only on English vision-language instruction pairs, it outperforms other VLMs on Korean benchmarks.}
\label{tab:cross-lingual-performance}
\end{table}

This zero-shot cross-lingual capability has significant implications for multilingual applications. While LLaVA models based on Vicuna \citep{vicuna2023} and Mistral \citep{jiang2023mistral7B} show limited Korean performance, Trillion-LLaVA excels despite identical English-only training. This suggests robust multilingual pre-training transfers effectively to multimodal tasks, potentially decoupling language-specific requirements from visual alignment training and streamlining multilingual vision-language system development.
These results indicate our model's multilingual foundation enables effective transfer of visual reasoning across languages without language-specific visual training data, raising questions about similar transfer effects for other languages and modalities.

\section{Conclusion, Limitations and Future Works}
We present Trillion-7B, demonstrating that effective multilingual capabilities can be achieved through architectural innovation rather than massive data scaling alone. Through our novel Cross-lingual Document Attention (XLDA) mechanism, strategic data filtering, and optimized tokenization strategy, we establish a new efficiency frontier for multilingual models, requiring just 59.4K H100 GPU hours while dedicating only 10\% of training tokens to non-English languages. The consistent performance across languages confirms XLDA's effectiveness at bridging knowledge between linguistic domains.

Our extensive experiments and ablation studies validate that properly designed cross-lingual training strategies significantly outperform traditional approaches that rely heavily on language-specific data scaling. These findings have important implications for democratizing access to powerful language models across diverse linguistic communities, as they suggest a path forward for developing high-performing multilingual models without the prohibitive computational and data requirements typically associated with frontier AI systems.

\paragraph{Limitations}
Our current model has several limitations. We dedicated limited resources to mathematical and coding data (less than 2\% of total training data) during pretraining, as our primary focus was on cross-lingual knowledge transfer. Consequently, Trillion-7B may exhibit sub-optimal performance on technical tasks. We plan to address this in the next generation of our models. Additionally, we performed minimal post-training optimization, and our safety mechanisms require further development to reach production standards. Users should note these limitations when deploying our models and must not use them to cause harm.

\paragraph{Future Works}
Future research directions include further improving multilingual model efficiency through enhanced training methods and architectural refinements. We plan to build upon our approach and enhance Trillion through dedicated post-training optimization and improved reasoning capabilities. Additionally, we aim to apply our techniques to vision models, as preliminary results already demonstrate the strength of our approach for cross-modal applications. Finally, we intend to scale up our methodology to develop a family of multilingual data-efficient models ranging from 70B-scale dense architectures to 400B-scale Mixture-of-Experts models, further democratizing access to powerful multilingual AI systems.

\bibliography{bib}

\newpage
\appendix

\section{Contribution}
\paragraph{Pretraining and Post-training}
 Suyeong An, Seungtaek Choi, Sungjun Han, Joanna Hong\footnote{All work done while at Trillion labs. Currently at Google Deepmind.}, Hyungguk Kim, Kyuseok Kim, Jamin Shin, Juyoung Suk, Wonsuk Yang 

\paragraph{Data Collection and Curation} Juneyoung Park, Yusik Kim

\paragraph{Infrastructure} Sunghoon Kang


\section{Baseline Models} \label{baseline_models}
Trillion-7B is compared against several widely recognized baseline models, all of which were accessed via the HuggingFace platform. Table~\ref{tab:baseline_models} summarizes each baseline model used in our experiments.

\begin{table}[ht]
\centering
\begin{tabular}{ll}
\toprule
\textbf{Model Name} & \textbf{Provider} \\ 
\midrule
\href{https://huggingface.co/trillionlabs/Trillion-7B-preview}{Trillion-7B-preview}& Our Model \\ 
\href{https://huggingface.co/LGAI-EXAONE/EXAONE-3.5-7.8B-Instruct}{EXAONE-3.5-7.8B-Instruct} & LG \\ 
\href{https://huggingface.co/google/gemma-2-9b-it}{Gemma-2-9B-it} & Google\\ 
\href{https://huggingface.co/meta-llama/Llama-3.1-8B-Instruct}{Llama-3.1-8B-Instruct} & Meta \\ 
\href{https://huggingface.co/Qwen/Qwen2.5-7B-Instruct}{Qwen2.5-7B-Instruct} & AliBaba \\ 
\href{https://huggingface.co/mistralai/Mistral-7B-Instruct-v0.3}{Mistral-7B-Instruct-v0.3} & Mistral AI \\ 
\bottomrule
\end{tabular}
\caption{Baseline models utilized for comparative evaluation. All models were accessed through the HuggingFace platform.}
\label{tab:baseline_models}
\end{table}

\newpage
\section{Full Evaluation Results and Details}
\label{app:full_details}

\paragraph{Full Evaluation Results} We detail full evaluation scores.

\begin{table}[ht]
    \centering
    \resizebox{\textwidth}{!}{
\begin{tabular}{@{}lrrrrrrr@{}}
\toprule
\textbf{Benchmark} & \textbf{Trillion-7B} & \textbf{EXAONE-3.5} & \textbf{Gemma-2-9b} & \textbf{Llama-3.1} & \textbf{Qwen2.5} & \textbf{SOLAR-10.7B} & \textbf{Mistral-7B} \\ \midrule
HellaSwag        & 58.94          & 60.04          & 59.72          & 59.81 & 61.97          & \textbf{68.72} & 65.79 \\
TruthfulQA\_mc1  & 36.10          & 40.64          & 42.96          & 38.07 & 47.74          & \textbf{56.18} & 42.47 \\
TruthfulQA\_mc2  & 54.10          & 59.74          & 60.09          & 54.54 & 64.72          & \textbf{70.64} & 59.41 \\
ARC:C            & 54.44          & 56.40          & \textbf{62.97} & 53.58 & 52.99          & 60.07          & 58.11 \\ \midrule
HAERAE           & \textbf{80.02} & 76.08          & 68.01          & 63.15 & 65.17          & 60.86          & 47.75 \\
KoBEST           & \textbf{79.61} & 78.57          & 79.98          & 70.09 & 79.24          & 75.20          & 66.50 \\ \midrule
KMMLU            & 48.09 & 45.39          & 46.66          & 41.41 & \textbf{50.15}         & 41.66          & 33.59 \\
MMLU             & 63.52          & 65.65          & 72.24          & 68.32 & \textbf{74.23} & 65.20          & 61.84 \\
GMMLU-en         & 67.75          & 69.50          & 76.25          & 67.50 & \textbf{77.25} & 71.75          & 65.50 \\
GMMLU-ko         & 60.75          & 60.00          & \textbf{64.25} & 54.00 & 59.25          & 53.75          & 43.00 \\
GMMLU-ja         & 60.75          & 45.75          & \textbf{66.50} & 54.50 & 65.75          & 50.75          & 50.00 \\
GMMLU-zh         & 59.50          & 50.00          & 63.75          & 60.25 & \textbf{68.75} & 57.00          & 47.25 \\
BBH              & 41.94          & 53.30          & 28.77          & 43.16 & \textbf{53.68} & 52.91          & 45.09 \\ \midrule
xwinograd\_en    & 87.78          & 87.10          & \textbf{89.55} & 88.09 & 85.63          & 87.35          & 88.39 \\
xwinograd\_jp    & 79.98 & 74.45          & \textbf{80.92}          & 76.02 & 72.89          & 72.58          & 70.70 \\
xwinograd\_zh    & 73.81          & 69.44          & 68.06          & 76.19 & \textbf{81.55} & 74.60          & 71.83 \\ \midrule
GPQA             & 32.81          & \textbf{38.61} & 36.83          & 30.58 & 34.15          & 28.35          & 32.59 \\ \midrule
\textbf{Average} & 61.17          & 60.63          & 62.79          & 58.78 & \textbf{64.42} & 61.62          & 55.87 \\ \bottomrule
\end{tabular}
    }
    \caption{General Reasoning and Knowledge}
    \label{tab:reasoning}
\end{table}

\begin{table}[ht]
    \centering
    \resizebox{\textwidth}{!}{
    \begin{tabular}{@{}lrrrrrrr@{}}
    \toprule
    \textbf{Benchmark} & \textbf{Trillion-7B} & \textbf{EXAONE-3.5} & \textbf{Gemma-2-9b} & \textbf{Llama-3.1} & \textbf{Qwen2.5} & \textbf{SOLAR-10.7B} & \textbf{Mistral-7B} \\ \midrule
    HumanEval        & 55.48 & 79.26          & 60.98 & 67.68 & \textbf{81.71} & 34.76 & 36.59 \\
    MBPP             & 40.40 & \textbf{61.40} & 8.40  & 39.20 & 51.00          & 29.40 & 36.00 \\ \midrule
    \textbf{Average} & 47.94 & 70.33          & 34.69 & 53.44 & 66.35          & 32.08 & 36.30 \\ \bottomrule
    \end{tabular}
}
    \caption{Coding}
    \label{tab:coding}
\end{table}

\begin{table}[ht]
    \centering
    \resizebox{\textwidth}{!}{
\begin{tabular}{@{}lrrrrrrr@{}}
\toprule
\textbf{Benchmark} & \textbf{Trillion-7B} & \textbf{EXAONE-3.5} & \textbf{Gemma-2-9b} & \textbf{Llama-3.1} & \textbf{Qwen2.5} & \textbf{SOLAR-10.7B} & \textbf{Mistral-7B} \\ \midrule
GSM8k            & 72.25 & 87.79 & 73.69 & 74.98 & \textbf{88.86} & 62.93 & 35.94 \\
MATH             & 32.70 & 70.68 & 41.06     & 38.30 & \textbf{71.50} & 14.38 & 12.12 \\
HRM8k            & 30.10 & 38.99 & 16.04 & 26.77     & \textbf{41.51} & 20.68 & 7.89  \\ \midrule
\textbf{Average} & 45.02 & 65.82 & 43.60 & 46.68 & \textbf{67.29} & 32.66 & 18.65 \\ \bottomrule
\end{tabular}
}
    \caption{Mathematical Reasoning}
    \label{tab:math}
\end{table}

\begin{table}[ht!]
    \centering
    \resizebox{\textwidth}{!}{
\begin{tabular}{@{}lrrrrrrr@{}}
\toprule
\textbf{Benchmark} & \textbf{Trillion-7B} & \textbf{EXAONE-3.5} & \textbf{Gemma-2-9b} & \textbf{Llama-3.1} & \textbf{Qwen2.5} & \textbf{SOLAR-10.7B} & \textbf{Mistral-7B} \\ \midrule
IFEval           & 79.13          & \textbf{81.42} & 75.48 & 74.93 & 75.85 & 51.61 & 52.64 \\
koIFEval         & \textbf{66.58} & 54.65          & 43.30 & 36.07 & 48.55 & 26.12 & 34.22 \\
MT-Bench (1-10)    & 7.00           & \textbf{8.15}  & 7.81  & 6.32  & 7.86  & 6.76  & 6.84  \\
KO-MT-Bench (1-10) & 6.27           & \textbf{8.13}  & 7.01  & 4.27  & 6.31  & 2.89  & 4.07  \\
LogicKor (1-10)    & 8.14           & \textbf{9.25}  & 8.33  & 6.45  & 7.99  & 1.85  & 4.76  \\ \midrule
\textbf{Average}   & 71.96          & \textbf{78.27} & 70.06 & 56.28 & 69.20 & 38.55 & 48.71 \\ \bottomrule
\end{tabular}
}
\caption{Instruction Following and Chat. For computing the average, the scores on the 0-10 scale are multiplied by 10.}
    \label{tab:instruction}
\end{table}

\paragraph{Prompts} \label{sec:eval_prompts}
We present the prompts used for evaluation reported in Table \ref{tab:performance-summary}. 

\begin{tcolorbox}[colback=black!10!white, 
colframe=black, title=\textbf{GSM8K/MATH prompt (CoT)}, fonttitle=\bfseries]

Q: \texttt{\{question\}}

Put your answer within \texttt{\textbackslash boxed\{\}}. Let's think step by step. 
\end{tcolorbox}
\noindent\begin{minipage}{\textwidth}
\captionsetup{justification=centering}
\captionof{figure}{Prompt for evaluating GSM8K (CoT) and MATH (CoT)}
\end{minipage}

\begin{tcolorbox}[colback=black!10!white, 
colframe=black, title=\textbf{MBPP prompt (CoT)}, fonttitle=\bfseries]

You are an expert Python programmer, and here is your task: \{text\} 

Your code should pass these tests:
\\

\{test\_list[0]\}

\{test\_list[1]\}

\{test\_list[2]\}
\\

Wrap your code in ` ` `python ` ` ` . Let's reason step by step.

\end{tcolorbox}
\noindent\begin{minipage}{\textwidth}
\captionsetup{justification=centering}
\captionof{figure}{Prompt for evaluating MBPP (CoT)}
\end{minipage}

\begin{tcolorbox}[colback=black!10!white, 
colframe=black, title=\textbf{Humaneval}, fonttitle=\bfseries]

You are an expert Python programmer, and here is your task: \{text\} 
\\

Wrap your code in ` ` `python ` ` ` . Let's reason step by step.

\end{tcolorbox}
\noindent\begin{minipage}{\textwidth}
\captionsetup{justification=centering}
\captionof{figure}{Prompt for evaluating Humaneval}
\end{minipage}

\paragraph{KoIFEval} To evaluate whether language models adhere effectively to instructions presented in Korean, we constructed the Korean Instruction Following benchmark (KoIFEval) based on the IFEval framework proposed in \citep{zhou2023instructionfollowingevaluationlargelanguage}. This benchmark dataset comprises 22 instruction types, selected from the original 25 of IFEval that are applicable to Korean. 

Each prompt in the benchmark includes one or more instructions and model performance is assessed at two levels: \textit{prompt-level}, which evaluates whether the generated response satisfies all included instructions, and \textit{instruction-level}, which assesses compliance with each instruction individually. The reported KoIFEval scores in this study represent the mean of these two metrics.

All data samples are generated using ChatGPT-4o, each conditioned on selected instructions. We verify whether the generated prompts appropriately include the selected instructions by employing a rule-based verification process.

\begin{tcolorbox}[colback=black!10!white, 
colframe=black, title=\textbf{KoIFEval}, fonttitle=\bfseries]

골든~스테이트~워리어스~선수들의~경기력~향상을~위한~스포츠~심리학적~접근~방안을~제시하세요. 두~개의~서로~다른~응답을~작성하세요.~응답은~반드시~6개의~별표로~구분해야합니다. 응답에~반드시~제목을~포함해야~하며,~제목은~<<기쁨의 시>>~와~같이~이중~꺾쇠~괄호로~감싸야~합니다. 

\end{tcolorbox}
\noindent
\begin{minipage}{\textwidth}
\captionsetup{justification=centering}
\captionof{figure}{Example of evaluating Ko-IFEval}
\end{minipage}

\begin{table}[!htbp]
    \centering
    \resizebox{\textwidth}{!}{%
\begin{tabular}{@{}lll@{}}
\toprule
\textbf{Instruction Group} &
  \textbf{Instruction} &
  \textbf{Description} \\ \midrule
Keywords &
  Include Keywords &
  응답에~반드시~\{keywords\}를~포함하세요. \\ \midrule
Keywords &
  Keyword Frequency &
  \begin{tabular}[c]{@{}l@{}}응답에서~단어~'\{keyword\}'이(가)~반드시 \\ 정확히~\{N\}번~\{relation\}~등장해야~합니다.\end{tabular} \\ \midrule
Keywords &
  Forbidden Words &
  응답에~\{forbidden\_words\}를~포함하지~마세요. \\ \midrule
Keywords &
  Letter Frequency &
  \begin{tabular}[c]{@{}l@{}}응답에서~문자~'\{letter\}'이(가)~반드시 \\ 정확히~\{N\}번~\{relation\}~등장해야~합니다.\end{tabular} \\ \midrule
Language &
  Response Language &
  \begin{tabular}[c]{@{}l@{}}응답~ 전체를 ~\{language\}로~ 작성하세요. \\ 다른~언어는~ 허용되지~ 않습니다.\end{tabular} \\ \midrule
Length Constraints &
  Number Paragraphs &
  \begin{tabular}[c]{@{}l@{}}응답에는 ~반드시~ {N}개의~ 단락이~ 포함되어야 ~합니다. \\ 단락들은 ~마크다운 ~구분선~ \texttt{* * *}으로 ~구분하세요.\end{tabular} \\ \midrule
Length Constraints &
  Number Words &
  응답을 ~\{relation\}~ \{N\}개~ 단어로~ 작성하세요. \\ \midrule
Length Constraints &
  Number Sentences &
  응답을~ \{N\}개 ~\{relation\} ~문장으로~ 작성하세요. \\ \midrule
Length Constraints &
  \begin{tabular}[c]{@{}l@{}}Number Paragraphs \\ + First Word in i-th Paragraph\end{tabular} &
  \begin{tabular}[c]{@{}l@{}}응답에는~ 반드시 ~\{N\}개의 ~단락이 ~있어야 ~하며, \\ 단락은 ~두 ~개의~ 줄~ 바꿈으로만 ~구분됩니다. \\ \{i\}번째 ~단락은 ~반드시~ \{first\_word\}로 ~시작해야~ 합니다.\end{tabular} \\ \midrule
Detectable Content &
  Postscript &
  \begin{tabular}[c]{@{}l@{}}응답의 ~마지막에~ 반드시~ \{postscript\_marker\}로 \\ 시작하는~ 첨언~ 추가하세요.\end{tabular} \\ \midrule
Detectable Content &
  Number Placeholder &
  \begin{tabular}[c]{@{}l@{}}응답에는~ 반드시 ~최소~ \{N\}개의~ 자리 ~\\ 표시자가 ~포함되어야 ~합니다.\end{tabular} \\ \midrule
Detectable Format &
  Number Bullets &
  \begin{tabular}[c]{@{}l@{}}응답에는 ~반드시 ~정확히 ~\{N\}개~ \{relation\} ~\\ 글머리 ~기호 ~목록이~ 포함되어야 ~합니다. \\ 마크다운 ~형식 ~(예: ~* ~이것은 ~항목입니다.)을 ~사용하세요.\end{tabular} \\ \midrule
Detectable Format &
  Title &
  \begin{tabular}[c]{@{}l@{}}응답에 ~반드시 ~제목을 ~포함해야~ 하며, \\ 제목은 ~<<기쁨의 ~시>>와 ~같이 ~\\ 이중~ 꺾쇠 ~괄호로~ 감싸야~ 합니다.\end{tabular} \\ \midrule
Detectable Format &
  Choose From &
  다음~ 옵션 ~중 ~하나로~ 응답하세요: \{options\} \\ \midrule
Detectable Format &
  Minimum Number Highlighted Section &
  \begin{tabular}[c]{@{}l@{}}응답에서~ 최소~ {N}개의~ 부분을 ~마크다운을 사용하여 \\ 강조하세요. ~예:~ \texttt{*강조된 부분*}.\end{tabular} \\ \midrule
Detectable Format &
  Multiple Sections &
  \begin{tabular}[c]{@{}l@{}}응답은 ~반드시~ \{N\}개의~ 섹션으로~ 구성되어야 ~합니다. \\ 각 ~섹션의~ 시작은~ \{section\_splitter\} \{번호\}. 로~ 표시하세요.\end{tabular} \\ \midrule
Detectable Format &
  JSON Format &
  출력 ~전체를~ JSON ~형식으로 ~감싸야~ 합니다. \\ \midrule
Combination &
  Repeat Prompt &
  \begin{tabular}[c]{@{}l@{}}먼저~ 요청을 ~변경 ~없이 ~반복한~ 후, ~응답을 작성하세요. \\ (요청을 ~반복하기~ 전에는 ~아무 ~말도 ~하지 ~마세요. \\ 반복해야~ 할~ 요청에는~ 이~ 문장이~ 포함되지 ~않습니다.)\end{tabular} \\ \midrule
Combination &
  Two Responses &
  \begin{tabular}[c]{@{}l@{}}두 ~개의 ~서로~ 다른~ 응답을~ 작성하세요. \\ 응답은~ 반드시~ 6개의~ 별표(\texttt{******})로~ 구분해야~ 합니다.\end{tabular} \\ \midrule
Start with / End with &
  End Checker &
  \begin{tabular}[c]{@{}l@{}}응답의 ~마지막을~ 반드시~ 이~ 정확한~ 문구~ \{end\_phrase\}로 끝내세요. \\ 이후에는~ 다른~ 단어가~ 따라오면~ 안~ 됩니다.\end{tabular} \\ \midrule
Start with / End with &
  Quotation &
  응답 ~전체를 ~큰따옴표로~ 감싸세요. \\ \midrule
Punctuation &
  No Commas &
  응답 ~전체에서 ~쉼표(,)를 ~사용하지~ 마세요. \\ \bottomrule
\end{tabular}
}
\caption{We selected 22 instructions from the 25 proposed in \citep{zhou2023instructionfollowingevaluationlargelanguage}, focusing on those that are adaptable to a Korean-centric context. Each instruction was refined with a Korean-oriented description, and the final set of 22 instructions was evenly distributed to ensure balanced coverage.}
\label{table:ko-ifeval-list}
\end{table}

\end{CJK*}
\end{document}